# Cascaded Multi-Granularity Pruning for On-Device LLM Inference in Industrial IoT

Jinghan Wang, *Student Member, IEEE* , Yanjun Chen, Wei Zhang, *Member, IEEE*, Xiaotong Huang, Tianchen Liu *Member, IEEE*, Gaoliang Peng*, *Member, IEEE*

***Abstract*—Deploying large language models (LLMs) on Industrial Internet of Things (IIoT) edge devices demands extreme compression, yet existing structured pruning methods collapse at high compression ratios due to one-shot importance estimation, and their cross-architecture behavior remains unpredictable. This article presents a cascaded multi-granularity pruning framework that removes layers, attention heads, and feed-forward channels in coarse-to-fine order, with lightweight low-rank recovery between stages to re-estimate component importance. An information-theoretic analysis motivates this ordering, and the Structural Independence Assumption (SIA) is formalized as a checkable condition predicting whether per-component pruning criteria are reliable for a given architecture: Multi-Head Attention (MHA)+GELU designs satisfy the SIA, whereas Grouped Query Attention (GQA)+SwiGLU designs violate it. On bearing fault diagnosis spanning 88M to 6.25B-parameter models, the framework extends achievable compression to 13.8× on MHA+GELU architectures with 83.82% accuracy (+3.70 percentage points (pp) over the strongest baseline), while exposing a ~74pp accuracy collapse on GQA+SwiGLU architectures that violate the SIA. Deployed on an industrial slewing bearing fault diagnosis platform with NVIDIA DGX Spark, compressed models reduce inference latency by up to 67.2% and peak memory by 62.5%, demonstrating viability for IIoT edge inference.**

## I. INTRODUCTION

INDUSTRIAL Internet of Things (IIoT) systems generate continuous high-frequency sensor streams that increasingly demand on-device intelligence. Uploading raw signals to a centralized cloud encounters three compounding bottlenecks: bandwidth that scales linearly with sensor density, round-trip latency incompatible with millisecond-level control loops, and privacy constraints that prohibit data from leaving the factory floor [1]. These factors have elevated edge intelligence, i.e., performing inference directly on resource-constrained nodes co-located with sensors, from an optimization option to a design imperative [2]. Large language models (LLMs), with their cross-task generalization capability, are now entering IoT perception workloads including predictive maintenance and fault detection [3], [4]. Even when memory suffices, a 6B-parameter FP16 model still occupies roughly 12 GB and incurs significant per-token compute, making real-time multi-model serving on a single node impractical in terms of latency and energy without compression. Structured pruning is therefore not an optimization nicety but the critical path for efficient on-device LLM inference in IIoT.

Among compression techniques, structured pruning directly reduces dense matrix dimensions, yielding speedups on any commodity hardware without sparse-tensor accelerators. Unstructured methods such as Wanda [5] and SparseGPT [6] achieve high sparsity but depend on specialized hardware support absent from most edge SoCs. Representative structured approaches include gradient-based LLM-Pruner [7], fluctuation-based FLAP [8], orthogonal-projection SliceGPT [9], and layer-removal methods ShortGPT [10] and LaCo [11]. At moderate compression (2-4×), these methods perform adequately. However, when pushed toward the compression ratios that edge deployment actually demands (>8×) and applied across the rapidly diversifying landscape of LLM architectures, three fundamental limitations emerge.

Existing structured methods operate at one or two granularity levels, typically attention heads and FFN channels, without exploiting layer-level redundancy. In our experimental setting, such single- or dual-granularity approaches reach at most approximately 9.4× compression, because the untouched granularity retains all its redundancy while the targeted granularity is over-pruned.

Current methods estimate component importance once, on the original dense model. Yet pruning itself alters the importance landscape: removing entire layers triggers importance redistribution among surviving components, causing their relative contributions to shift substantially. This discrepancy grows with the compression ratio, explaining why one-shot methods collapse beyond moderate compression; capturing redistribution inherently requires interleaving pruning with recovery and re-estimation, which no single-pass criterion can achieve.

The same criterion can produce negligible accuracy loss on one architecture yet catastrophic collapse on another at comparable compression, and existing observations of such sensitivity [9], [12] remain empirical with no predictive power. For heterogeneous edge fleets, such trial-and-error behavior is

This work was supported by the National Natural Science Foundation of China (No. 52275099). Corresponding author: Gaoliang Peng (pgl7782@hit.edu.cn).

Jinghan Wang, Xiaotong Huang, Tianchen Liu, and Gaoliang Peng are with the School of Mechanical and Electrical Engineering, Harbin Institute of Technology, Harbin 150001, China (e-mail: 23B908011@stu.hit.edu.cn; 25S108110@stu.hit.edu.cn; tianchenliu@hit.edu.cn; pgl7782@hit.edu.cn).

Yanjun Chen and Wei Zhang are with the School of Information Science and Technology, Eastern Institute of Technology, Ningbo, China (e-mail: yychen@eitech.edu.cn; zhw@eitech.edu.cn).

disqualifying: practitioners need an a priori applicability criterion before committing to a pruning strategy.

These three limitations reflect a deeper problem: structured pruning research remains almost entirely empirically driven, lacking (i) an information-flow analysis prescribing how to order multi-granularity removal to minimize information loss, and (ii) validity conditions delineating the architectures on which independent importance estimation is reliable. This work addresses how to organize pruning, why coarse-to-fine is optimal, and when the framework is trustworthy.

To address these questions, this paper proposes a theory-grounded multi-granularity cascaded pruning framework, validated on industrial bearing fault diagnosis (a representative IIoT edge workload imposing near-maximal domain shift from pretraining) across 88M-6.25B models with end-to-end deployment on NVIDIA DGX Spark. The main contributions are summarized as follows.

1) A multi-granularity cascaded pruning framework that interleaves coarse-to-fine structural removal (layers, heads, FFN channels) with staged LoRA recovery, capturing the importance redistribution that one-shot methods miss.
2) An information-theoretic analysis that models the LLM as a Markov processing chain and applies the Data Processing Inequality (DPI) to motivate, under stated assumptions, the coarse-to-fine ordering, with per-granularity heuristic loss bounds linked to the Layer Contribution Ratio.
3) A formalization for the LLM structured pruning setting, of the Structural Independence Assumption (SIA) underlying per-component pruning criteria, with analysis showing that Multi-Head Attention (MHA)+Gaussian Error Linear Unit (GELU) architectures satisfy it while Grouped Query Attention (GQA)+Swish-Gated Linear Unit (SwiGLU) architectures violate it, yielding a checkable a priori applicability criterion.
4) End-to-end deployment validation on an industrial slewing bearing test rig equipped with NVIDIA DGX Spark across models spanning two orders of magnitude in scale, demonstrating practical IIoT edge deployability.

The rest of the paper is organized as follows. Section II reviews related work; Section III presents the framework and its theoretical analyses; Section IV reports experimental and deployment results; Section V concludes the paper.

## II. Related Work

### A. On-Device LLMs and Edge Intelligence in IoT

The integration of LLMs into IoT systems is an active research direction [1]. On the systems side, inference engines such as llama.cpp enable on-device execution through kernel-level optimization, while collaborative frameworks such as EdgeShard [2] partition models across distributed edge devices. Post-training quantization, including GPTQ [13] and AWQ [14], reduces bit-width orthogonally to pruning, and combining both yields multiplicative gains. Split inference, however, still depends on network connectivity, violating IIoT privacy and availability constraints. Meanwhile, LLM-based industrial fault diagnosis methods [3], [4] have demonstrated promising accuracy but report results exclusively on workstation GPUs without addressing edge deployability. Existing on-device efforts thus optimize the inference stack or bit-width, leaving parameter count as the dominant memory bottleneck, which structured pruning attacks directly.

### B. Structured Pruning of Large Language Models

Magnitude-based pruning has a long history [15], [16]. In the LLM era, unstructured methods, including Wanda [5] and SparseGPT [6], achieve high sparsity but require sparse-tensor hardware absent from edge SoCs. Structured approaches form the primary edge pathway. LLM-Pruner [7] ranks grouped weights by gradient sensitivity. FLAP [8] introduces a fluctuation-based criterion at head and FFN granularities. SliceGPT [9] reduces embedding dimensions via orthogonal projection. At the layer granularity, ShortGPT [10] and LaCo [11] exploit near-identity Transformer blocks, but layer-only pruning is itself a single-granularity strategy. All of these methods estimate importance once on the dense model at one or two granularities, neither exploiting layer-level redundancy jointly with finer granularities nor capturing pruning-induced importance redistribution, which causes their collapse at high compression ratios (Section IV-C).

### C. Multi-Granularity and Iterative Pruning with Recovery

Iterative prune-retrain cycles [15] were abandoned in the LLM era due to the prohibitive cost of full fine-tuning. LoRA [17] has revived lightweight recovery, spawning methods such as LoRAPrune [18] and LoRAShear [19], although both apply recovery as a single terminal step. AMP [12] ablates attention-versus-MLP pruning on LLaMA-2, observing severe head sensitivity but without theoretical explanation. VTrans [20] also proposes multi-granularity pruning, but its granularity ordering is heuristic, recovery serves only as post-processing without feeding into importance re-estimation, and no open-source code is available. None of these methods interleaves recovery with importance re-estimation across granularities, nor explains theoretically why any particular order should be preferred and this is the question our information-theoretic analysis addresses.

### D. Pruning Theory and Architecture Sensitivity

Theoretical pruning work has focused on existence results: the Lottery Ticket Hypothesis [21] proves that sparse sub-networks exist but does not address criterion effectiveness. The Information Bottleneck framework [22] applies information-theoretic analysis to deep networks, providing the conceptual precedent for our DPI-based modeling [23]. On the empirical side, FLAP [8] reports KV head groups far more sensitive than FFN modules on LLaMA, and AMP [12] documents a 23.39 percentage points (pp) collapse from head-only pruning on LLaMA-2. These observations remain architecture-specific post-hoc descriptions with no predictive power. Moreover, Grouped Query Attention [24] and SwiGLU [25] are now defaults in LLaMA, Mistral, Qwen, and ChatGLM, making violating architectures the mainstream. To our knowledge, no prior work explicitly formalizes independence conditions for LLM per-component pruning criteria and our Structural Independence Assumption (Definition 1) provides this missing condition, explains the above observations as special cases, and yields a checkable, predictive criterion for unseen architectures.

## III. METHODOLOGY

### A. Problem Formulation and Framework Overview

The dense model pruned in this work is the three-phase LLM-based bearing fault diagnosis framework of [26], comprising knowledge-guided text generation (ChatGLM), time-series feature extraction (GPT-2), and multi-modal fusion (GPT-2). Given a pre-trained LLM and a target task dataset, the structured pruning problem is formulated as:

$$\min_{\theta'} \mathcal{L}(f_{\theta'}, \mathcal{D}) \text{ s.t. } |\theta'| \le \gamma \cdot |\theta|, \tag{1}$$

where $\theta$ denotes the full parameter set of the original model, $\theta'$ denotes the parameter set after pruning, $\mathcal{D}$ is the target task dataset, $\gamma \in (0,1)$ is the target retention ratio (i.e., the compression ratio is $1/\gamma$), and $\mathcal{L}$ is the task loss. Rather than treating $\theta$ as a flat parameter vector, we decompose it along three structural granularity levels inherent to the Transformer architecture:

$$\theta = \{\Theta_{layer}, \Theta_{head}, \Theta_{ffn}\}, \tag{2}$$

where $\Theta_{layer}$ denotes complete Transformer blocks (coarse granularity), $\Theta_{head}$ denotes individual attention heads (medium granularity), and $\Theta_{ffn}$ denotes FFN intermediate dimensions (fine granularity). These three are the only structural units removable without breaking the cross-layer residual interface, and all three must be included because FFN alone accounts for approximately two-thirds of each block's parameters ($8d^2$ vs $4d^2$), so omitting any granularity caps achievable compression. This decomposition enables a cascaded pruning pipeline that operates from coarse to fine: layer pruning first reduces representational depth, head pruning then refines the attention structure within surviving layers, and FFN pruning finally adjusts per-layer transformation capacity. Between each stage, LoRA-based staged recovery stabilizes the model before the next pruning granularity is evaluated. The complete pipeline is illustrated in Fig. 1 and formalized in Algorithm 1.

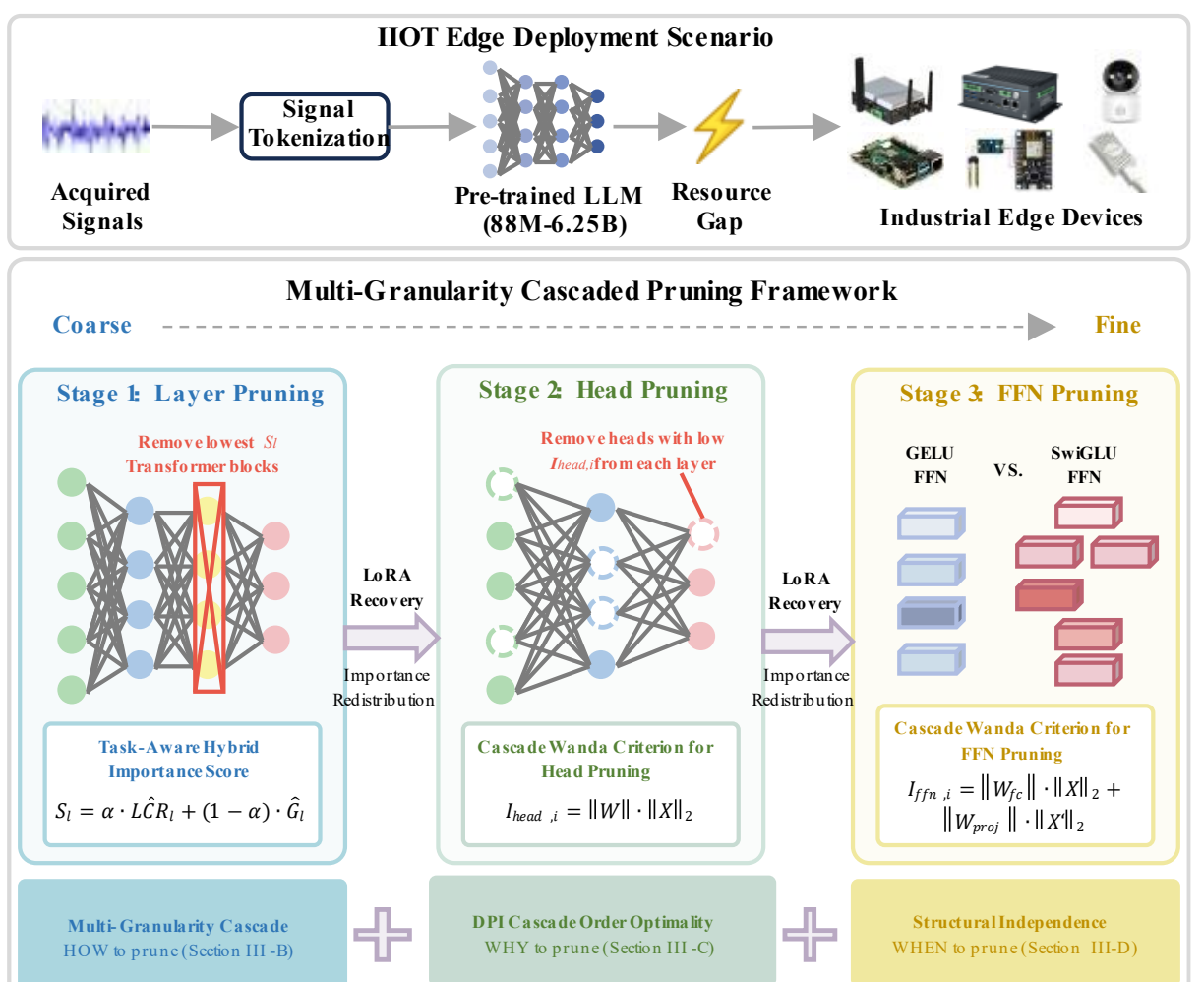

**Fig. 1.** Overview of the proposed multi-granularity cascaded pruning framework for IIoT edge deployment. The pipeline proceeds from coarse to fine granularity, including layer pruning, attention head pruning, and FFN channel pruning with LoRA-based staged recovery and importance redistribution between consecutive stages. The three theoretical contributions respectively address how to prune across granularities, why the coarse-to-fine ordering is optimal, and when the cascaded independence assumption holds.

### B. Multi-Granularity Cascaded Pruning Framework

#### 1) Task-Aware Layer Importance Estimation

At the coarsest granularity, the proposed framework removes entire Transformer blocks based on a hybrid importance score that combines structural contribution with task-specific relevance. The structural component is the Layer Contribution Ratio (LCR), defined as:

$$LCR_l = \mathbb{E}_{x\sim\mathcal{D}}\left[\frac{\left\|h_l^{out} - h_l^{in}\right\|_2}{\left\|h_l^{in}\right\|_2}\right], \tag{3}$$

where $x \sim \mathcal{D}$ denotes an input sample drawn from the dataset, and $h_l^{in}$, $h_l^{out}$ are the input and output hidden states of layer $l$ produced by the forward pass of $x$ through the network. A low LCR indicates that the layer performs a near-identity transformation and can be removed with minimal representational loss. However, LCR alone is task-agnostic; we therefore introduce a gradient-based task-aware component:

$$G_l = \sum_{p\in\theta_l} \left\| p \odot \nabla_p \mathcal{L} \right\|_1, \tag{4}$$

which measures the sensitivity of the task loss to each layer's parameters. $\odot$ denotes the Hadamard (element-wise) product, so $p \odot \nabla_p \mathcal{L}$ multiplies each parameter value by its corresponding gradient, measuring per-parameter sensitivity to the task loss. The final task-aware importance score is:

$$S_l = \alpha \cdot L\hat{C}R_l + (1-\alpha) \cdot \hat{G}_l, \tag{5}$$

where $L\hat{C}R_l$ and $\hat{G}_l$ denote min-max normalized forms of the raw scores $LCR_l$ and $G_l$, and $\alpha = 0.5$ balances the two components. Layers with the lowest $S_l$ are removed first, with the first and last layers protected as the information entry and exit points. The computational cost is one forward pass (for LCR) plus one backward pass (for gradients).

#### 2) Cascade Wanda Criterion for Head and FFN Pruning

At the medium and fine granularity levels, attention heads and FFN channels are pruned using the Wanda criterion [5], which evaluates component importance by the product of weight magnitude and input activation norm:

$$I_{wanda}(\omega) = |\omega| \cdot \|X\|_2, \tag{6}$$

where $X$ denotes the input activation matrix received by a given layer during the forward pass of calibration samples $x \sim \mathcal{D}$. Specifically, the importance of head $i$ in layer $l$ is defined as:

$$I_{head,i}^{(l)} = \sum_{W\in\{W_Q^i, W_K^i, W_V^i\}} \sum_j \left|W_j\right| \cdot \left\|X_j\right\|_2. \tag{7}$$

Heads are ranked independently within each layer, with the constraint that at least one head is retained per layer. For FFN pruning, the importance of intermediate dimension $i$ is defined as:

$$I_{ffn,i} = \sum_j \left|W_{fc(i,j)}\right| \cdot \left\|X_j\right\|_2 + \sum_j W_{proj(j,i)} \cdot \left\|X'_j\right\|_2, \tag{8}$$

where $X$ is the input activation to the up-projection. FFN dimensions are ranked globally across all layers $W_{fc}$ and $X' = \text{GELU}(XW_{fc})$ is the input activation to the down-projection $W_{proj}$.

A critical distinction is the cascade aspect of importance evaluation: head importance is computed on the model that has already undergone layer pruning and Low-Rank Adaptation (LoRA) recovery, and FFN importance after both preceding stages, whereas one-shot methods evaluate all components on the original model. As established in Section III-C, staged re-estimation yields more accurate rankings because the remaining components have already redistributed their roles.

*3) Staged LoRA Recovery*

To recover the information lost by structural removal at each cascade stage, LoRA is applied to the pruned model:

$$\Delta W = \boldsymbol{BA},\ \boldsymbol{B} \in \mathbb{R}^{d \times r}, \boldsymbol{A} \in \mathbb{R}^{r \times d},\ r \ll d\ , \quad (9)$$

where only the low-rank matrices $\boldsymbol{B}$ and $\boldsymbol{A}$ are updated while the original weights remain frozen. After training, the adaptation is merged ($\boldsymbol{W}' = \boldsymbol{W} + \Delta W$) to incur no additional inference cost.

The staged recovery protocol consists of three phases: 10-epoch LoRA recovery after layer pruning, 10-epoch recovery after head pruning, and 30-epoch recovery after FFN pruning. This design serves two functions beyond accuracy restoration. First, it enables importance redistribution: after stage-*k* pruning and recovery $\mathcal{R}_k$, the remaining components absorb the functions of pruned components, causing their importance rankings to shift. The importance estimation at stage *k+1* therefore reflects the true post-pruning structure rather than the original model (formalized as Proposition 1 in Section III-C). Second, the low-rank constraint acts as implicit regularization, particularly effective under the cross-load domain shift (Section IV-D).

*4) Complete Algorithm*

Algorithm 1 summarizes the complete multi-granularity cascaded pruning pipeline, integrating the three pruning stages with their respective importance criteria and staged LoRA recovery phases.

The total computational cost comprises three lightweight LoRA training phases (less than 1% of model parameters per phase) and three forward passes for importance estimation, substantially lower than full fine-tuning.

**Algorithm 1** Multi-Granularity Cascaded Pruning with Staged Recovery

**Require:** Pre-trained model $f_\theta$, dataset $\mathcal{D}$, targets $(n_{\text{layer}}, n_{\text{head}}, n_{\text{ffn}})$, LoRA rank $r$
**Ensure:** Compressed model $f_{\theta'}$
// Stage 1: Layer Pruning (Coarse)
1: Compute task-aware scores $S_l = \alpha \cdot \text{LCR}_l + (1-\alpha) \cdot G_l$ for all layers
2: Protect first and last layers
3: Remove lowest-scoring intermediate layers $\rightarrow f_{\theta_1}$
4: LoRA recovery $(r,\ 10 \text{ epochs}) \rightarrow$ merge $\rightarrow f_{\theta'_1}$
// Stage 2: Head Pruning (Medium) — on recovered model
5: Collect activations on $f_{\theta'_1}$; compute Cascade Wanda head importance
6: Per-layer ranking; remove lowest heads (keep $\geq 1$ per layer) $\rightarrow f_{\theta_2}$
7: LoRA recovery $(r,\ 10 \text{ epochs}) \rightarrow$ merge $\rightarrow f_{\theta'_2}$
// Stage 3: FFN Pruning (Fine) — on recovered model
8: Collect activations on $f_{\theta'_2}$; compute Cascade Wanda FFN importance
9: Global ranking; remove lowest FFN dimensions $\rightarrow f_{\theta_3}$
10: LoRA recovery $(r,\ 30 \text{ epochs}) \rightarrow f_{\theta'}$
11: **return** $f_{\theta'}$

*C. Information-Theoretic Analysis of Cascade Ordering*

The coarse-to-fine ordering of Section III-B admits a principled motivation: modeling the LLM as a Markov information processing chain, we apply the Data Processing Inequality (DPI)to provide an information-theoretic motivation under the following three assumptions. Note that this analysis tracks $I$ ($X$; model output) rather than the task-specific $I$ (Y; representation); while the Information Bottleneck framework optimizes the latter, our use of DPI serves as a heuristic motivation for the coarse-to-fine ordering rather than a tight optimality guarantee.

Assumption A1 (Stochastic processing chain): The model input is a noisy observation $X = S + \varepsilon$, where $S$ is the source signal and $\varepsilon$ represents sensor noise inherent to IIoT vibration acquisition, ensuring that the chain $X \rightarrow h_1 \rightarrow h_2 \cdots \rightarrow h_L \rightarrow Y$ is a genuinely stochastic Markov chain and that all mutual information quantities are non-degenerate.

Assumption A2 (Within-granularity consistency): Within each granularity level, the per-component importance criterion is assumed to provide a rank-preserving approximation of component importance within each layer. This assumption is motivated by, but not formally implied by, the structural independence analysis in Section III-D, which establishes intra-layer computational independence rather than global scoring fidelity.

Assumption A3 (Smoothness): Each layer mapping $h_l = g_l(h_{l-1})$ is Lipschitz continuous, ensuring that bounded input perturbations produce bounded output perturbations.

*1) LLM as Markov Information Processing Chain*

$$X \rightarrow h_1 \rightarrow h_2 \cdots \rightarrow h_L \rightarrow Y\ , \quad (10)$$

where $X$ is the input representation, $h_L$ is the hidden state after layer $l$, and $Y$ is the model output. Each layer is an information processing node, forming a Markov chain. By the DPI [23], for any Markov chain $X \rightarrow Y \rightarrow Z$:

$$I(X;Z) \leq I(X;Y)\ . \quad (11)$$

That is, information can only be lost, never created, through processing.

*2) Cascade Order Motivation*

Building on the Markov chain formulation, the following proposition motivates the coarse-to-fine cascade ordering by combining the DPI with a pruning budget efficiency argument.

Proposition 1 (Coarse-to-Fine Cascade Motivation): For sufficiently high compression ratios (empirically $1/\gamma > 4\times$), let $\prod_{layer}$, $\prod_{head}$, $\prod_{ffn}$ denote pruning operations at three granularity levels, each followed by staged recovery $\mathcal{R}$. For any target compression ratio $1/\gamma$, the coarse-to-fine ordering is:

$$\prod_{layer} \rightarrow \mathcal{R}_1 \rightarrow \prod_{head} \rightarrow \mathcal{R}_2 \rightarrow \prod_{ffn} \rightarrow \mathcal{R}_3\ , \quad (12)$$

which satisfies $I(X; f_{\theta_{c2f}}(X)) \geq I(X; f_{\theta_\sigma})$ for a generic alternative ordering $\sigma$ at the same total compression, where $\theta_{c2f}$ and $\theta_\sigma$ denote the pruned parameters under the respective orderings.

Removing layer $l$ eliminates node $h_l$ from the Markov chain, connecting $h_{l-1}$ directly to $h_{l+1}$ via a skip connection. By the continuity of mutual information, layers with low LCR incur minimal information loss, as:

$$I(X; h_{l+1}) \approx I(X; h_{l-1})\ , \quad (13)$$

when $h_l^{out} \approx h_l^{in}$. Crucially, layer removal preserves the internal computation of surviving layers, keeping their head and FFN structures intact. After layer pruning and recovery $\mathcal{R}_1$, head importance computed on the surviving model $f_{\theta'_1}$ is conditioned on the finalized layer structure, yielding a more accurate ranking than evaluation on the original $f_\theta$ because the search space is reduced and the remaining heads have redistributed their functions through $\mathcal{R}_1$. The same argument applies to FFN importance on $f_{\theta'_2}$.

Conversely, under the reverse ordering $\prod_{ffn} \rightarrow \prod_{head} \rightarrow \prod_{layer}$, FFN dimensions pruned at the first stage may reside in layers subsequently removed by $\prod_{layer}$, wasting pruning budget. Let $P_{waste}$ denote this fraction: under coarse-to-fine ordering $P_{waste} = 0$ by construction, whereas any ordering where coarse pruning follows fine pruning yields $P_{waste} > 0$. Combining these arguments, the coarse-to-fine ordering is expected to better preserve $I(X; f_{\theta'}(X))$ since each stage operates on finalized structure and no budget is wasted.

*3) Per-Granularity Information Loss Bound*

Beyond ordering optimality, it is important to quantify the information loss from each individual pruning operation, directly linking the LCR metric to information-theoretic guarantees.

Proposition 2 (Heuristic Information Loss Bound): Let layer $l$ have $LCR_l = \rho_l$. The mutual information loss after removing layer $l$ satisfies:

$$I(X;Y) - I(X;Y_{-l}) \le \rho_l \cdot H(h_{l-1}), \tag{14}$$

where $Y_{-l}$ denotes the output with layer $l$ removed and $H(h_{l-1})$ is the entropy of the layer input.

Removing layer $l$ replaces $h_l^{out}$ with $h_l^{in}$, introducing a perturbation $\delta_l = h_l^{out} - h_l^{in}$ whose relative magnitude is precisely $LCR_l$ as defined in (3). Applying the bound on mutual information loss under bounded perturbation [27] yields $\Delta I \le \rho_l \cdot H(h_{l-1})$.This bound is conservative and may not be tight in practice, as the right-hand side involves the full hidden-state entropy which can greatly exceed the task-relevant information; nevertheless, it qualitatively supports LCR-based ranking: layers with smaller LCR admit tighter information loss bounds and can be pruned with greater confidence.

*4) Importance Redistribution under Staged Recovery*

The cascade framework re-estimates importance at each stage rather than relying on a single one-shot evaluation, which raises the question of whether rankings change after recovery.

Proposition 3 (Importance Redistribution): Let $\mathrm{Im}_k(c_j)$ denote the importance of component $c_j$ estimated at cascade stage $k$. After pruning at stage $k$ and staged recovery $\mathcal{R}_k$, the importance satisfies:

$$\mathrm{Im}_{k+1}(c_j) \ne \mathrm{Im}_k(c_j), \tag{15}$$

for remaining components $c_j$ that is, the importance ranking of surviving components is not invariant under staged recovery.

This follows from the fact that LoRA recovery modifies model weights $W' = W + \Delta W$, altering activation statistics and the gradient landscape. Components previously redundant may become critical after neighboring components are removed and the model adapts. Consequently, stage-($k$+1) importance estimation better reflects the post-pruning structure than one-shot evaluation on the original model, an effect amplified at higher compression ratios, as validated in Section IV-D.

*D. Structural Independence Analysis*

The cascade framework relies on per-component importance evaluation, implicitly assuming that each head or FFN channel can be scored in isolation. The following analysis formalizes this assumption and proves the architectural conditions under which it holds or fails.

*1) Structural Independence Assumption*

Definition 1: A neural network architecture satisfies the SIA for a per-component importance criterion $\mathcal{C}$ if, for any two structural components $c_i$, $c_j (i \ne j)$ is:

$$P(c_j(x) \mid \mathrm{prune}(c_i)) \approx P(c_j(x)) \text{ for any input } x, \tag{16}$$

that is, removing $c_i$ does not substantially alter the computed function of $c_j$.

This definition is implicitly shared by all per-component criteria in the literature, including L1-norm, Wanda $(|W| \cdot \|X\|_2)$, and activation-based metrics. When violated, these scores may systematically misrepresent the impact of component removal.

*2) MHA+GELU Satisfies Independence*

Proposition 4 (MHA+GELU Independence): In Transformer architectures using MHA with independent Q, K, V projections and GELU-activated FFN, the SIA is satisfied at both head and FFN channel granularities.

In MHA, the output is:

$$\mathrm{Multihead}(x) = \mathrm{Concat}(\mathrm{head}_1, \dots \mathrm{head}_H) W_O, \tag{17}$$

where $\mathrm{head}_i$ is formulated as:

$$\mathrm{head}_i = \mathrm{Softmax}\left(\frac{xW_Q^i (xW_K^i)^T}{\sqrt{d_k}}\right) xW_V^i, \tag{18}$$

Each head possesses independent projections $W_Q^i, W_K^i, W_V^i$, removing head $i$ eliminates its rows in the concatenation and corresponding columns in $W_O$ but does not alter any other head within the same layer $j$'s computation, satisfying Definition 1. For the FFN, GELU is element-wise, so in $\mathrm{FFN}(x) = \mathrm{GELU}(xW_1)W_2$, channel $j$'s activation depends exclusively on *j-th* row of $W_1$ and *j-th* column of $W_2$, removing channel $j$ does not affect any other channel.

*3) GQA+SwiGLU Violates Independence*

Proposition 5 (GQA+SwiGLU Violation): In Transformer architectures using Grouped Query Attention and SwiGLU-activated FFN, the SIA is violated at both head and FFN channel granularities.

In GQA, $n_q$ query heads share $n_{kv}$ key-value heads, each serving $D_{GQA} = n_q / n_{kv}$ query heads. Removing one KV head simultaneously disables $D_{GQA}$ dependent query heads. For ChatGLM-2 with $n_q = 32$ and $n_{kv} = 2$, meaning a single KV head removal affects 16 query heads, violating Definition 1 with dependency degree $D_{GQA}$. For SwiGLU, the computation $\mathrm{SwiGLU}(x) = \mathrm{Swish}(xW_{gate}) \odot (xW_{up})$ creates a paired constraint: dimension *j*'s output depends jointly on *j-th* row of $W_{gate}$ and $W_{up}$. Per-component criteria evaluate these independently and may retain one while removing the other, breaking the pair and forcing $\mathrm{output}_j = 0$ regardless of the retained channel's magnitude.

*4) Applicability Boundary*

The preceding propositions delineate a clear applicability boundary, synthesized as follows. Per-component structured pruning criteria provide reliable importance ranking when the target architecture satisfies the SIA. For violating architectures, group-aware criteria that respect structural dependencies are required.

This yields a two-step verification procedure for new architectures: (1) check whether attention heads share key-value projections (MHA: $D$=1, independent; GQA/MQA: $D$>1, dependent), and (2) check whether FFN activations involve paired computations (GELU/ReLU: independent; SwiGLU/GeGLU: dependent). Section IV-E validates this empirically: the same L1-norm criterion at comparable compression produces a 74pp accuracy divergence between the two architecture families.

## IV. Experimental Evaluation

### A. Experimental Setup

*1) Dataset and Task Definition*

We evaluate our framework on the Case Western Reserve University (CWRU) bearing fault diagnosis benchmark [28] , which is the most widely adopted standard dataset in rotating machinery health monitoring. The dataset comprises vibration signals from four operating conditions, namely Normal (N), Inner race fault (I), Outer race fault (O), and Ball fault (B), collected under four motor load settings (0–3 HP).

To rigorously assess generalization rather than memorization, we adopt a cross-load evaluation protocol in which models are trained on data from 0, 1, and 2 HP operating conditions and tested exclusively on the 3 HP condition, which is entirely unseen during training. This simulates realistic operating-condition shift and imposes a train-test distribution gap that penalizes overfitting (cf. Section IV-D).

*2) Model Configurations*

Four LLM-based fault diagnosis models spanning two orders of magnitude in parameter count are used, covering two architectural families, as listed in Table I.

TABLE I
Configurations Used in This Study

| Case | Model | Architecture | Parameters (Million) |
|---|---|---|---|
| **Phase 1** | ChatGLM-2 6B | GQA + SwiGLU | 6,247 |
| **Phase 2** | GPT-2 | MHA + GELU | 88 |
| **Fusion** | GPT-2 | MHA + GELU | 134 |
| **FD-MVLLM** [4] | GPT-2 | MHA + GELU | 177 |

The GPT-2-based models employ standard Multi-Head Attention with independent head parameters and element-wise GELU activation, while ChatGLM-2 uses GQA (with 2 KV-head groups shared by 32 query heads, $D_{GQA}$=16) and SwiGLU activation with paired gate-up projections. This architectural contrast provides natural conditions for the sensitivity analysis in Section IV-E.

*3) Comparison Methods*

Four representative pruning methods are selected for comparison, as summarized in Table II. Magnitude Pruning uses the same L1-norm criterion as our framework but without cascade ordering or staged recovery, serving as a direct ablation baseline. Since FLAP, LLM-Pruner, and Wanda are designed for LLaMA-scale (7B+) models, we re-implement each method's core importance criterion and adapt it to structured (head/FFN) granularity on our models under identical data and compression targets. This ensures criterion-level comparison but may not reflect each method's full effectiveness on its original target architecture. All methods use the same LoRA-based recovery protocol to isolate the effect of the pruning criterion. Different compression ratios are achieved by jointly varying the number of retained layers, attention heads per layer, and FFN intermediate dimensions as specified in Algorithm 1.

TABLE II
Comparison Methods and Their Core Pruning Criteria

| Method | Venue | Core Criterion | Type |
|---|---|---|---|
| **FLAP** [8] | AAAI 2024 | Var(activation) × L2(weight) | Structured, training-free |
| **LLM-Pruner** [7] | NeurIPS 2023 | Gradient × weight group importance | Structured, gradient-based |
| **Wanda** [5] | ICLR 2024 | ‖W‖ × ‖X‖ (weight × activation) | Unstructured, training-free |
| **Magnitude** | Baseline | Pure L1-norm, one-shot | Structured, training-free |

*4) Implementation Details*

For the cascaded pruning pipeline, layer importance is measured by the task-aware hybrid score $S_l = \alpha \cdot L\hat{C}R_l + (1-\alpha)\cdot \hat{G}_l (\alpha = 0.5)$ ,combining the LCR from a single forward pass. Head and FFN channel importance are evaluated using the Cascade Wanda criterion $I_{wanda}(\omega) = |\omega| \cdot \|X\|_2$ , which jointly considers weight magnitude and input activation norm. The first and last transformer layers are protected from removal in all experiments.

We employ LoRA for post-pruning knowledge recovery with rank 1/*r*=4 and scaling factor *s*=16, applied to the attention projection layers ($c_{attn}$, $c_{proj}$ for GPT-2 models; $qkv_{proj}$ and dense for ChatGLM-2). In the improved cascade pipeline, staged recovery is applied after each pruning granularity, consisting of 10-epoch recovery after layer and head pruning, respectively, followed by 30-epoch full recovery after FFN pruning. Although LoRA adapts only attention projections, FFN pruning damage is indirectly compensated because the adapted attention outputs serve as inputs to the FFN sublayers, allowing the network to redistribute representational capacity followed by recovery after FFN pruning. The optimizer is AdamW with a learning rate of $5\times10^{-4}$, cosine annealing schedule (minimum lr $10^{-6}$), weight decay of 0.01, and batch size of 16.

Training and pruning experiments are conducted on an NVIDIA GeForce RTX 4090 GPU, while edge deployment validation uses an NVIDIA DGX Spark 128GB module. All experiments are repeated with three random seeds (42, 123, 456), and we report mean and standard deviation unless otherwise specified. The software environment consists of PyTorch 2.6.0 with CUDA 12.4, Hugging Face Transformers, and the PEFT library for LoRA integration.

### *B. Overall Framework Effectiveness*

Table III reports the overall results across all four model configurations, and the framework achieves consistent compression across two orders of magnitude in model scale (88M to 6.25B), with accuracy degradation not exceeding 3.00% in all cases. Notably, FD-MVLLM exhibits a +2.09% accuracy improvement after pruning, which can be attributed to the implicit regularization effect of removing redundant parameters from an over-parameterized model (177M for a 4-class task). Layer pruning eliminates the most structurally redundant blocks identified by LCR, removing the primary source of over-parameterization while preserving the task-critical hierarchy.

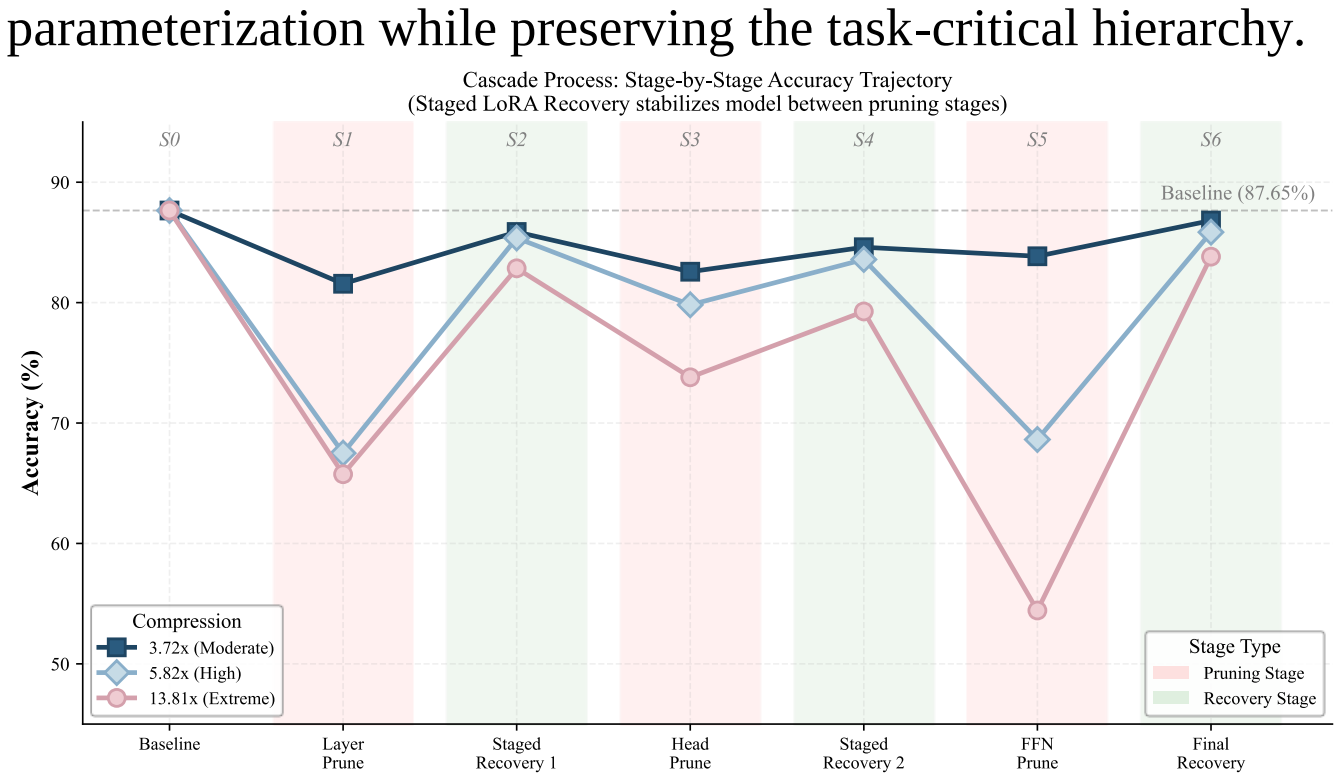

**Fig. 2.** Stage-by-stage accuracy trajectory of the cascaded pruning process at three compression levels (3.72×, 5.82×, 13.81×). The sawtooth pattern reflects alternating pruning (accuracy drop) and staged LoRA recovery (accuracy restoration) phases.

TABLE III
OVERALL PERFORMANCE OF THE PROPOSED FRAMEWORK ACROSS FOUR LLM-BASED FAULT DIAGNOSIS MODELS

| Case | Original Parameters (Million) | Pruned Parameters (Million) | Compression Ratio | Accuracy Change |
|---|---|---|---|---|
| **Phase 1** | 6,247 | 2,326 | 2.69× | −3.00% |
| **Phase 2** | 88 | 61 | 1.44× | −1.05% |
| **Fusion** | 134 | 76 | 1.75× | −0.08% |
| **FD-MVLLM** | 177 | 128 | 1.38× | +2.09% |

Fig. 2 visualizes the cascaded pruning process at three compression levels (3.72×, 5.82×, and 13.81×). A characteristic sawtooth trajectory is observed, where each pruning step causes an accuracy drop followed by a recovery phase. At 13.81× compression, layer pruning causes a steep decline from 87.65% to approximately 69.55%, yet staged recovery restores the model sufficiently for subsequent importance estimation, ultimately achieving 83.82% final accuracy. The sawtooth amplitude increases with the compression ratio, confirming that staged recovery becomes increasingly essential at high compression. This is precisely why importance estimation at stage $k$+1 (after recovery $\mathcal{R}_k$ ) is more accurate than a one-shot estimation on the original model, as formalized in Proposition 3.

To assess cross-dataset generalizability, the framework is further evaluated on the Jiangnan University (JNU) bearing dataset [29] (baseline: 79.82±4.83%) at three compression levels, as shown in Table IV. The results mirror the regime-dependent pattern observed on CWRU: at moderate compression (5.88×), one-shot methods achieve higher accuracy, while at 13.94× the cascade framework achieves 71.63%, outperforming Wanda by 4.22pp and Magnitude by 7.41pp, confirming the cross-dataset validity of the importance redistribution mechanism.

TABLE IV
CROSS-DATASET VALIDATION IN ACCURACY ON JNU DATASET

| Compression | Ours | Wanda | Magnitude |
|---|---|---|---|
| **2.50×** | 76.76±4.27 | 75.73±5.36 | 75.86±4.90 |
| **5.88×** | 73.99±5.69 | 74.13±4.55 | 74.10±4.26 |
| **13.94×** | 71.63±4.20 | 67.41±4.46 | 64.22±5.24 |

### *C. Comparison with State-of-the-Art Methods*

We compare our framework against four representative methods across six compression levels (1.3×–13.8×) on the Phase 2 model (GPT-2, 88M parameters), with results summarized in Table V and visualized as box plots in Fig. 3. All results report 3-seed mean and standard deviation. Note that in our re-implementation, FLAP and Wanda are configured without layer pruning and therefore do not reach compression ratios beyond approximately 9.4× and their entries at 13.8× are marked as unavailable.

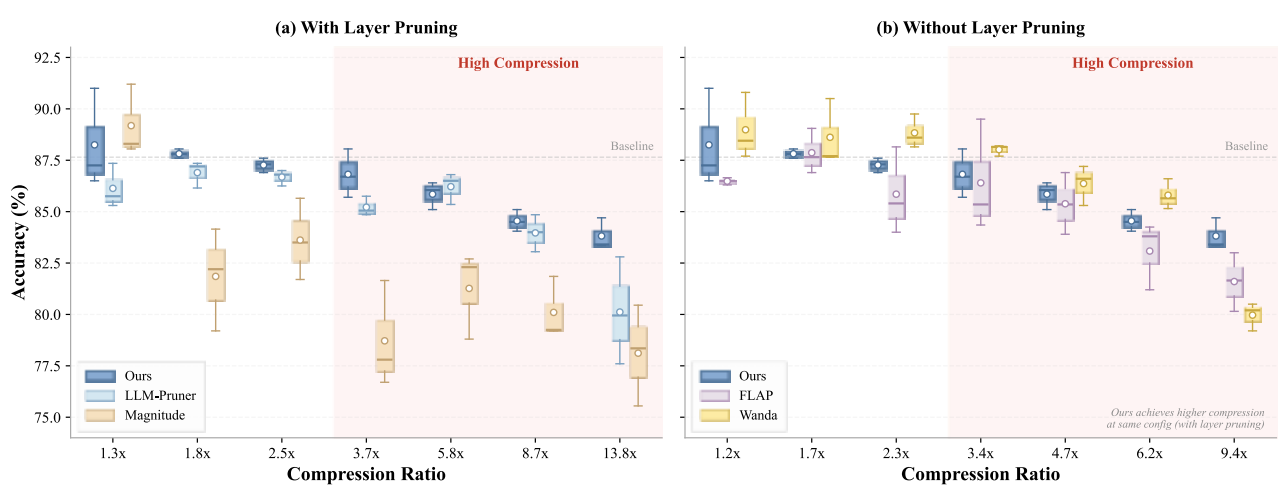

**Fig. 3.** Accuracy-compression comparison of five pruning methods. (a) Methods supporting layer pruning. (b) Methods without layer pruning (FLAP, Wanda).

The results reveal a regime-dependent performance landscape. At low compression (1.3×–3.7×), simpler one-shot criteria such as Wanda and Magnitude outperform the proposed method. This is expected from a theoretical perspective, because when few components are removed, importance redistribution across the remaining structure is negligible, and the overhead of cascade coordination offers no benefit over direct global ranking. Practical edge deployment, however, demands precisely the high-compression regime where one-shot methods fail.

At compression ratios beyond approximately 5.8×, a clear crossover occurs. At 13.8×, the proposed method achieves 83.82%, leading LLM-Pruner by 3.70pp and Magnitude by 5.70pp, while Wanda and FLAP cannot reach this regime at all due to the lack of layer pruning support. This growing advantage follows directly from the importance redistribution mechanism (Section III-C): the cascade re-estimates importance after each structural change and recovery, whereas one-shot methods rely on rankings computed on the intact model. Fig. 3 confirms this trend, with the most gradual accuracy decline beyond 5.8×.

TABLE V
ACCURACY COMPARISON WITH STATE-OF-THE-ART PRUNING METHODS

| Compression Ratio | Ours | Wanda | LLM-Pruner | FLAP | Magnitude |
|---|---|---|---|---|---|
| **~1.3×** | 88.25±1.97 | 88.98±1.32 | 86.13±0.88 | 86.48±0.14 | 89.18±1.43 |
| **~2.5×** | 88.27±0.29 | 88.83±0.67 | 86.68±0.32 | 85.85±1.72 | 83.62±1.61 |
| **~3.7×** | 87.82±0.96 | 88.02±0.22 | 85.22±0.39 | 86.40±2.23 | 78.72±2.12 |
| **~5.8×** | 85.85±0.55 | 85.80±0.60 | 86.22±0.62 | 85.38±1.22 | 81.27±1.75 |
| **~8.7×** | 84.55±0.43 | 79.97±0.56 | 83.97±0.74 | 81.60±1.16 | 80.10±1.24 |
| **~13.8×** | 83.82±0.62 | - | 80.12±2.13 | - | 78.12±2.01 |

Fig. 4 visualizes this difference through head importance heatmaps, where each cell encodes the L1 importance of a head (row) per layer (column). The cascade produces sharper, sparser maps because it evaluates heads after the layer structure has stabilized. The cascade evaluates head importance after the layer structure has stabilized, whereas one-shot methods may preserve heads that appear individually important but reside in low-contribution layers, yielding suboptimal retention at high compression.

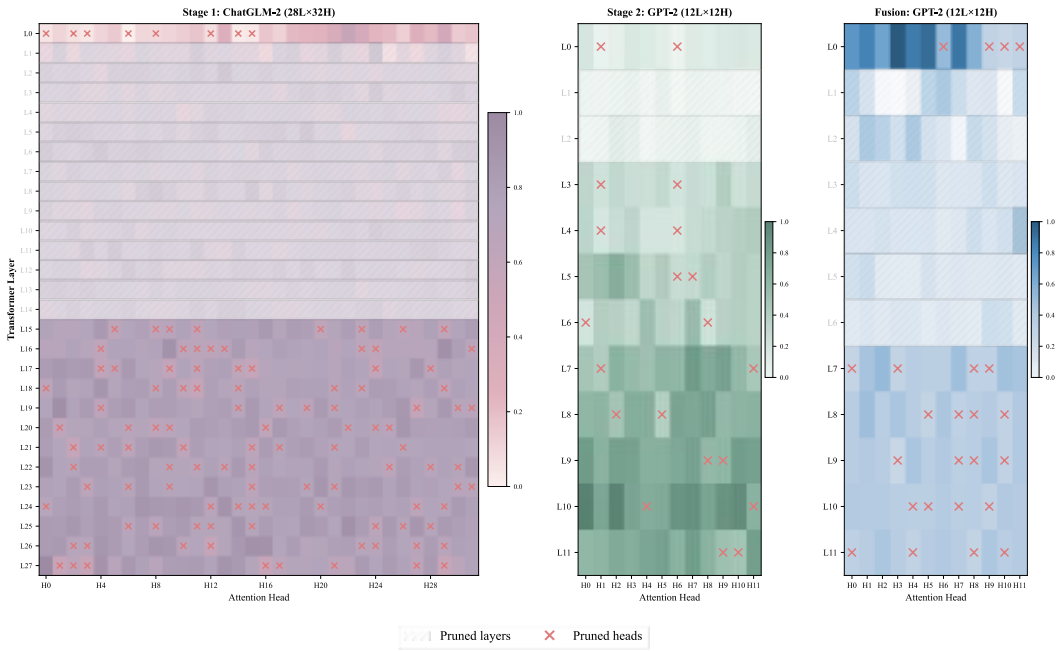

**Fig. 4.** Head importance distribution and pruning decisions across three model stages. Color intensity indicates normalized head importance; × markers denote pruned heads; hatched regions denote pruned layers.

### D. Ablation Studies

#### 1) Cascade Order Ablation

We test the ordering motivated in Section III-C (L→H→F optimal) on the Phase 2 model by comparing four orderings at three compression levels, with results shown in Fig. 5.

Sensitivity to cascade order grows monotonically with compression: the best–worst gap expands from 1.78pp at 1.82× to 10.36pp at 8.67×, consistent with the DPI-motivated importance redistribution mechanism.

At low compression (1.82×), Simultaneous is best (83.98%) because inter-component interactions are negligible. This reverses at higher compression: Simultaneous drops to 72.00% at 8.67× (vs. 74.08% for L→H→F), since without staged recovery it evaluates all components on a model whose rankings no longer reflect the post-pruning structure. The reverse order F→H→L collapses to 63.72%, substantially degraded, because FFN channels pruned first may have been critical for attention heads evaluated in the subsequent stage, corrupting the head importance signal and causing cascading misestimation.

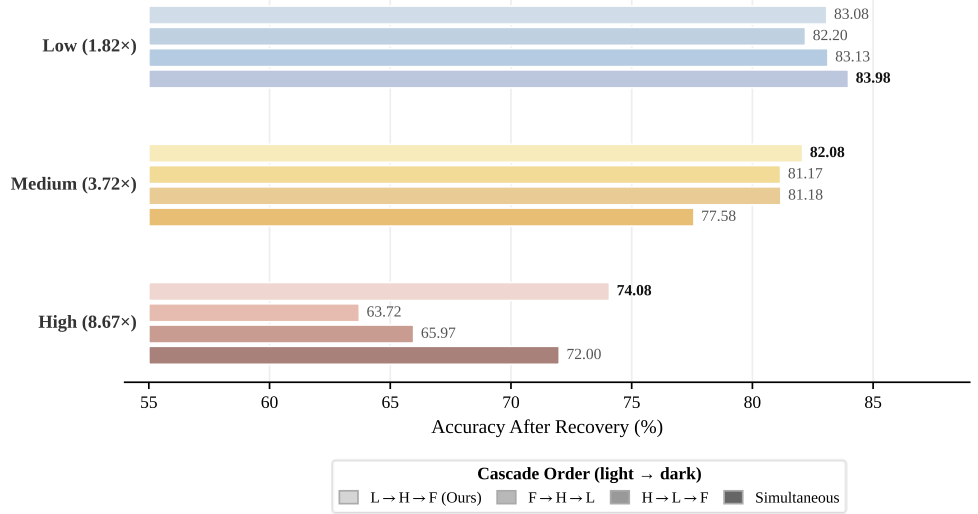

**Fig. 5.** Cascade order ablation across three compression levels. Four orderings are compared: L→H→F (proposed), F→H→L (reverse), H→L→F, and Simultaneous.

#### 2) Per-Granularity Contribution Ablation

We ablate individual granularity levels on the Fusion model with the others fixed, plus a Layer-Only baseline; results are in the upper panel of Fig. 6, aggregate accuracy and compression results are reported below, with per-class recall and F1 breakdowns. The results reveal complementary roles across granularity levels. Layer pruning dominates accuracy preservation (Layer-Only achieves 95.08%, within 0.27pp of Full), but alone provides only 1.47× compression. Head and FFN pruning supply the additional 0.28× to reach 1.75×. This reflects the hierarchical nature of transformer redundancy: near-identity layers (low LCR) can be removed wholesale, while head- and channel-level redundancy is distributed within the remaining layers. Removing any single granularity degrades accuracy, compression, or both.

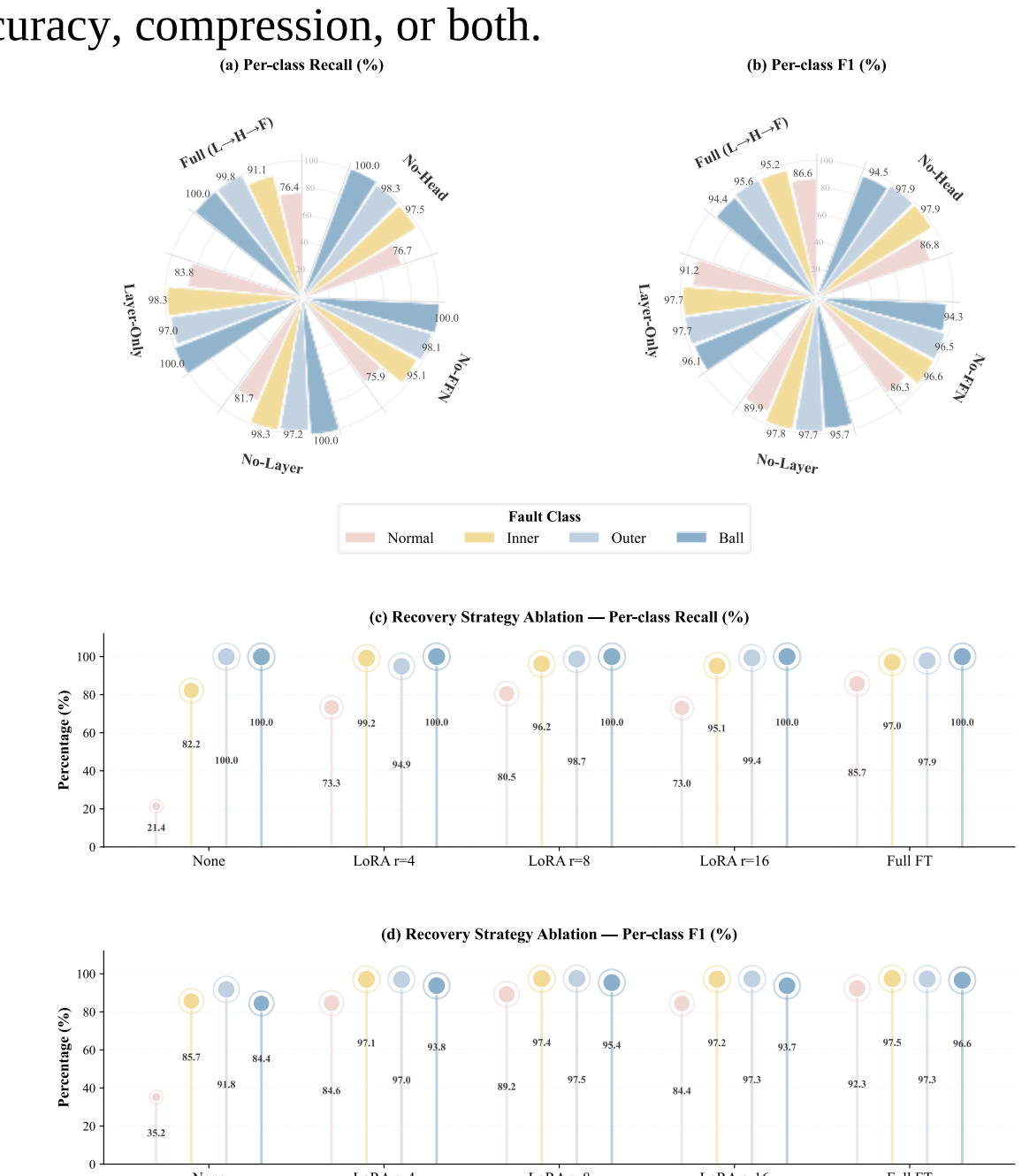

**Fig. 6.** Ablation study results. (a)–(b) Per-granularity contribution ablation: per-class recall and F1-score under five configurations (Full, No-Layer, No-Head, No-FFN, Layer-Only). (c)–(d) Recovery strategy ablation: per-class recall and F1-score for five recovery strategies (None, LoRA r=4/8/16, Full FT).

#### 3) Recovery Strategy Ablation

To validate LoRA-based recovery and determine the optimal rank, we compare five configurations visualized in the lower panel of Fig. 6. The 18.05 pp gap between no recovery and LoRA $r$=4 confirms that post-pruning adaptation is essential.

LoRA $r$=4 (96.17%) outperforms full fine-tuning (94.13%) despite using only 11.58% of trainable parameters, with accuracy decreasing monotonically as rank increases (r=4>r=8>r=16>Full FT). Under the cross-load protocol, the distribution shift between training (0–2 HP) and testing (3 HP) conditions penalizes over-parameterized recovery. The low-rank constraint acts as implicit regularization, forcing the model to learn compact, generalizable recovery directions that suffice to restore the few key representational subspaces disrupted by pruning.

In summary, the ablations confirm that cascade order is critical, with L→H→F optimal at meaningful compression (supporting Section III-C); that all three granularity levels contribute complementary improvements; and that LoRA-based staged recovery is superior to unconstrained fine-tuning under domain shift.t.

### E. Architecture Sensitivity Analysis

To empirically examine the SIA, the Fusion model (GPT-2, MHA+GELU) and Phase 1 model (ChatGLM-2, GQA+SwiGLU) are pruned at comparable compression ratios (2.03× and 1.95×) using the same L1-norm criterion, with all other variables controlled to the extent possible (Table VI). We acknowledge that the two models differ in scale and baseline accuracy (94.69% vs. 88.75%), so the observed divergence may partly reflect optimization difficulty at larger scale in addition to the architectural factor.

TABLE VI

ARCHITECTURE SENSITIVITY CONTROLLED EXPERIMENT

| Metric | GPT-2 (MHA+GELU) | ChatGLM-2 (GQA+SwiGLU) |
|---|---|---|
| **Baseline accuracy** | 94.69% | 88.75% |
| **Post-pruning accuracy** | 94.13% | 19.75% |
| **After FT recovery** | 95.16% | 87.25% |

**Note**: same L1-norm criterion at comparable compression ratios.

As shown in Table VI, an ~74pp divergence in post-pruning accuracy (94.13% vs. 19.75%) constitutes the central empirical evidence for the SIA. The collapse of ChatGLM-2 is consistent with two compounding mechanisms. In GQA, removing a single KV head disables 16 query heads simultaneously ($D_{GQA}$ =16), but L1-norm reflects only the KV head's own parameters, systematically underestimating its true contribution by a factor proportional to $D_{GQA}$. In SwiGLU FFN, L1-norm independently ranks gate and up channels, potentially breaking the paired constraint Swish $(xW_{gate}) \odot (xW_{up})$ and forcing unpaired dimensions to zero.

The recovery to 87.25% after fine-tuning suggests a criterion failure rather than a capacity problem: fine-tuning re-learns the coupling relationships, bypassing the independence assumption. The LCR distributions in Fig. 7 further illustrate this pattern, as GPT-2 exhibits well-separated layer importance scores while ChatGLM-2's distribution is distorted by shared KV dependencies. Fig. 8 illustrates the structural contrast between MHA's independent head projections and GQA's dependency fan-out. These results confirm the predictive value of the two-step verification procedure of Section III-D.

### F. Edge Deployment Evaluation

To evaluate deployment potential, inference experiments were conducted on an industrial slewing bearing with an NVIDIA DGX Spark 128GB. As visualized in the lower right corner of Fig. 9, all pruned models reduce latency (up to 67.2%), memory footprint (up to 62.5%), and energy consumption. The most pronounced gains appear in Phase 1 and the full cascade, where peak VRAM drops by over 61%. Fusion achieves the highest energy efficiency improvement (+154.9%), as throughput more than doubles while power consumption remains nearly unchanged. These results confirm the end-to-end viability of the proposed approach for LLM-based industrial IoT applications.

## V. CONCLUSION

This paper presented a multi-granularity cascaded pruning framework with staged LoRA recovery for on-device LLM inference in industrial IoT. Interleaving coarse-to-fine removal with lightweight adaptation captures the importance redistribution that one-shot methods inherently miss. An information-theoretic analysis via the DPI justifies the coarse-to-fine ordering under stated assumptions, while the formalized SIA provides a checkable criterion for predicting, prior to pruning, whether L1-type criteria are reliable for a given architecture. Experiments on industrial bearing fault diagnosis across 88M-6.25B models show that the framework achieves superior accuracy retention at high compression ratios and extends achievable compression to 13.8×. In practice, the framework is most beneficial beyond 4× compression and applies directly to architectures satisfying structural independence (e.g., MHA+GELU), while the Structural Independence analysis explains ~74pp accuracy divergence on GQA+SwiGLU designs that require group-aware criteria. Deployment evaluation on an industrial slewing bearing platform with NVIDIA DGX Spark shows up to 67.2% latency reduction and 62.5% memory saving, demonstrating practical potential for IIoT edge inference.

A current limitation is that the empirical validation covers a single industrial scenario and one edge-class GPU; whether the observed compression–accuracy trade-offs generalize to other IoT verticals and more resource-constrained devices remains to be verified. Future work includes validation on additional industrial benchmarks and deployment platforms, combining pruning with quantization, and developing group-aware criteria for violating architectures.

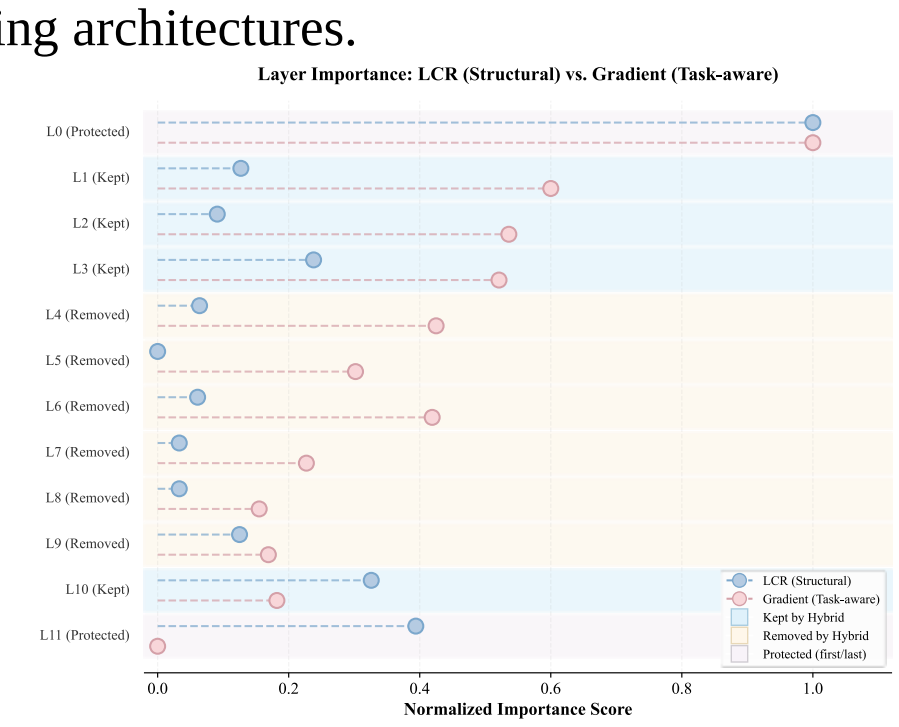

**Fig. 7.** Layer importance analysis comparing structural (LCR) and task-aware (gradient-based) importance scores across all transformer layers. Blue/red

shading indicates layers retained or removed by the hybrid criterion. The first and last layers are protected.

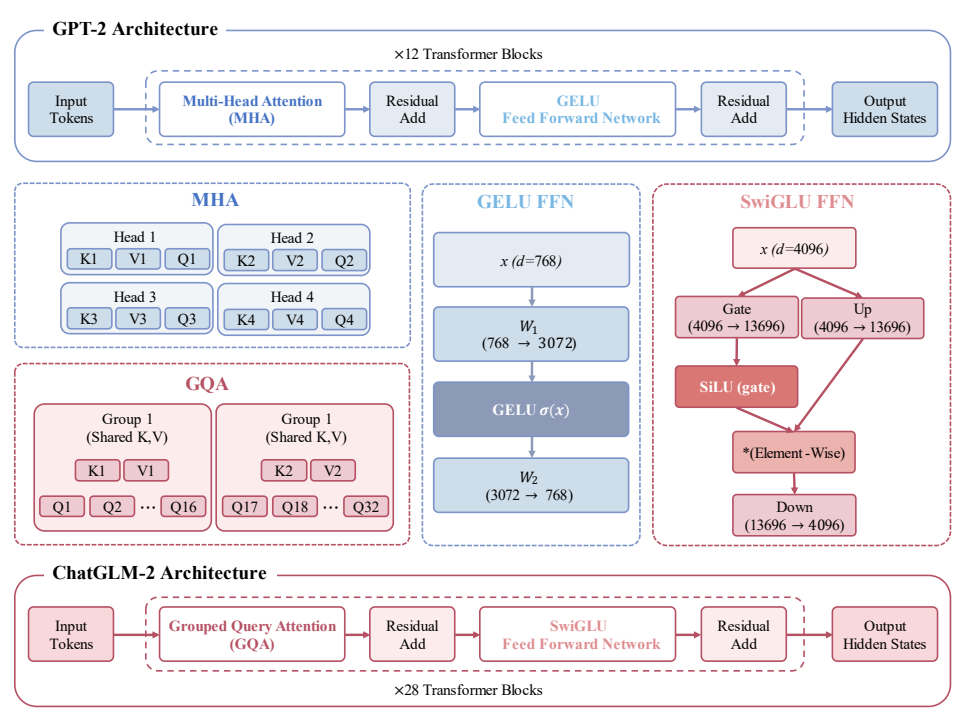

**Fig. 8.** Architectural comparison between GPT-2 and ChatGLM-2, including full transformer block diagrams, attention mechanism and FFN contrasts.

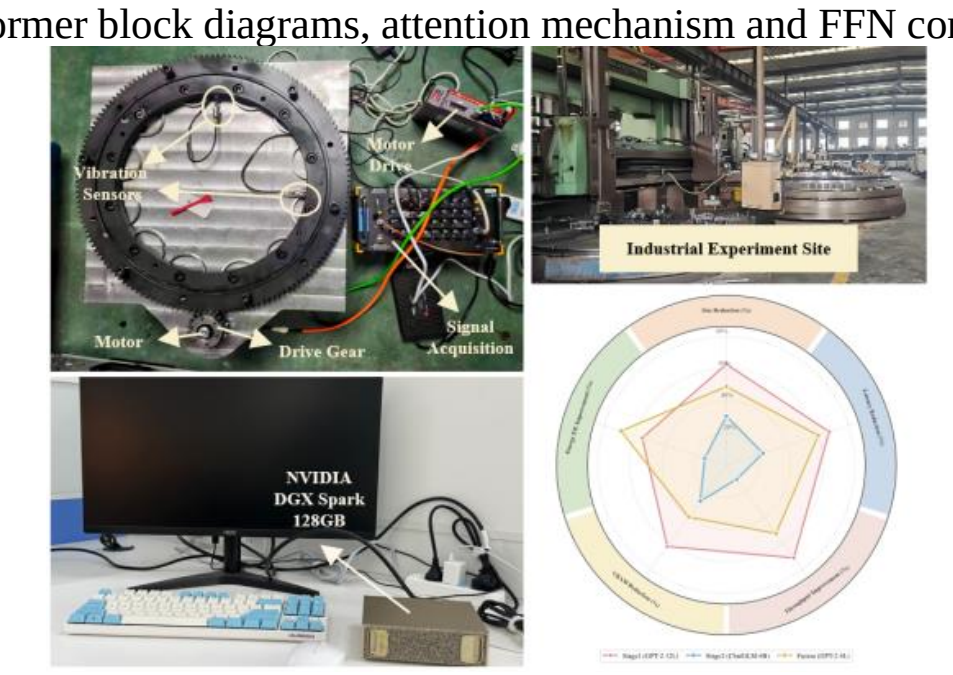

**Fig. 9.** Industrial experimental setup and the corresponding edge deployment efficiency profile on NVIDIA DGX Spark 128GB. Five metrics are compared across three pruned models: size reduction, latency reduction, throughput improvement, VRAM reduction, and energy efficiency improvement.